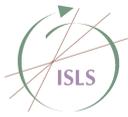

# Science and Engineering for What? A Large-scale Analysis of Students' Projects in Science Fairs

Adelmo Eloy, Universidade de São Paulo, adelmo.eloy@usp.br
Thomas Palmeira Ferraz, Télécom Paris, thomas.palmeira@telecom-paris.fr
Fellip Silva Alves, Universidade de São Paulo, fellipalves15@usp.br
Roseli de Deus Lopes, Universidade de São Paulo,roseli.lopes@usp.br

**Abstract:** Science and Engineering fairs offer K-12 students opportunities to engage with authentic STEM practices. Particularly, students are given the chance to experience authentic and open inquiry processes, by defining which themes, questions and approaches will guide their scientific endeavors. In this study, we analyzed data from over 5,000 projects presented at a nationwide science fair in Brazil over the past 20 years using topic modeling to identify the main topics that have driven students' inquiry and design. Our analysis identified a broad range of topics being explored, with significant variations over time, region, and school setting. We argue those results and proposed methodology can not only support further research in the context of science fairs, but also inform instruction and design of contexts-specific resources to support students in open inquiry experiences in different settings.

## Introduction

Participating in Science and Engineering Fairs (SEFs) provide students with an opportunity to showcase the outcomes of inquiry processes, developing a better understanding of science and an increased interest in STEM fields. (e.g., Grinnell et al., 2020). SEFs also offer students a space for authentic participation in "doing science," by engaging with science and engineering practices, as well as taking a sense of commitment and ownership over their projects (e.g., Koomen et al., 2018). Similarly, SEFs are often used to illustrate how students can experience open inquiry, since they can define themes, questions, and approaches in a continuous decision-making process (Zion & Mendelovici, 2012). In particular, while students are able to bring their passions and interests into the scientific inquiry process (Adler et al., 2018), with limited research on how such agency in problem finding manifests in SEFs (LaBanca, 2012). To help address this gap, we propose the following research questions:

1. What are the main topics students participating in SEF choose to explore in their projects?
2. How do those topics vary in time, school settings, and region where they come from?

In this study, we employed topic modeling, a machine learning technique, to analyze data from over 5,000 projects presented in a nationwide science and engineering fair in Brazil. Our analysis revealed the main themes driving students' investigation and design in their projects, and how they are responsive to their context, illustrating how students' interests manifest in their work. We argue these findings can inform designers and teachers on what kind of resources would be most relevant to support students in open inquiry processes.

## Methods

### Setting: The Brazilian Science and Engineering Fair (FEBRACE)

FEBRACE is a major outreach program of the Universidade de São Paulo launched in 2003 that annually holds the largest nationwide SEF in Brazil, where students from all states showcase hundreds of projects after being selected from over 2,000 submissions per year. To join FEBRACE, teams up to three students from public and private schools submit their projects for evaluation by a jury or are selected by one of the associated regional fairs. The submission materials include a paper/report and a five-minute video presenting their project. Selected students present their project in a synchronous event and undergo a further round of evaluation and feedback. FEBRACE publishes information about the projects selected (e.g., title, authors, institution, abstract and keywords) in annual proceedings available on their website. Outstanding projects are then selected for the International Science and Engineering Fair (ISEF), to which FEBRACE is affiliated.

### Data collection and analysis

In this study, we collected information from 5,296 projects accepted and presented at FEBRACE, from 2003 to 2022. Information of each projected included title, keywords, abstract, year presented, school setting, and state/region of origin. FEBRACE proceedings were the main data sources for the dataset (additional information was provided by the organizing committee), which is available upon request.

We used topic modeling to identify major topics among students' projects. Topic modeling is a statistical model that consists in extracting topical patterns within a collection of documents (Egger & Yu, 2022). It has been widely applied to educational research, such as identifying the relationship between the topic relevance of pre-service teachers' journals and their grades (Chen et al., 2016), and identifying major topic from students' essay to support the design of culturally adaptive learning experiences (Coelho & McCollum, 2021).

More specifically, we used the BERTopic (Grootendorst, 2022), which is a deep learning-based model that takes a set of documents, clusters it into topics, and generates representative words for each topic. Before using the BERTopic, we employed text pre-processing techniques, namely *Stopword Removal* (cutting non-significant parts of the vocabulary) and *Lemmatization* (converting nouns and adjectives to their masculine and singular form, and verbs to their infinitive form) (Ferraz et al., 2021).

Two automatic metrics guided the BERTopic: *coherence* (measures an average of the degree of semantic similarity between the words that represent each topic) and *diversity* (measures the percentage of unique words that represents the topics, which means how varied the whole set of topics is) (Dieng et al., 2020). BERTopic managed to assign 58% of the projects in 72 topics with at least 10 projects, which is a reasonable performance given the challenges of automatic categorization (Alcoforado et al., 2022). We manually proposed shorter terms based on the representative words of each topic and reviewed them with external professionals from the FEBRACE organizing committee.

We compared the distribution of topics across three variables: year (grouped into 4-year intervals), region (corresponding to Brazil's 5 macro regions) and school setting (categorized as public or private, which were the most prevalent in our sample). We used the Chi-square statistical test of independence to determine if there were statistically significant differences between groups for each variable. Finally, we identified topics with the highest values of dispersion and described top-5 lists for each variable to expand the findings from the statistical test.

## Findings

Using topic modeling, we classified 3,087 projects from our sample into 72 topics (42.9 projects per topic, SD=36.9). Topic coherence was 0.62, which aligns with typical results in the literature (Röder et al., 2015), and topic diversity was 0.72, indicating that the topics are distinguishable from each other (Dieng et al., 2020). Figure 1 shows the distribution of projects across the topics, with the top 10 topics (colored in yellow) containing 1,188 or 38.5% of the projects, while the bottom half of topics (colored in dark gray, each with 30 or fewer projects) represent 765 or 24.8% of the projects. A complete list of topics is available at bit.ly/isls-2023-sef-topics.

**Figure 1**.
*Number of occurrences for each topic identified in the model, together with a general top-10 list of topics.*

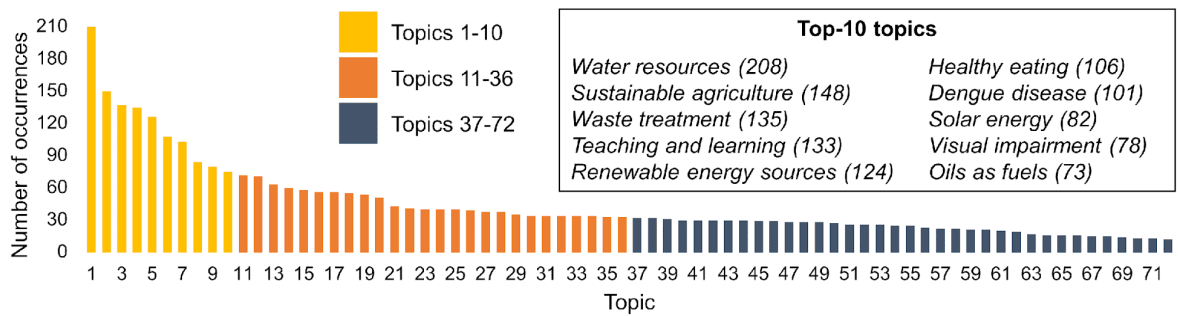

Table 1 presents results from the Chi-square Test of Independence on the topics across three variables: *year* (grouped in 5 four-year ranges, from 2003-2006 to 2019-2022), *region* (corresponding to Brazil's 5 macro regions), and *school setting* (public or private schools). Degrees of freedom (DoF) were adjusted for variables with 5 categories due to the requirement of at least 5 projects in each category, resulting in fewer topics (N_topics). Despite this, p-values ($p<0.05$) and strength of association ($0.1<\omega<0.3$) indicated the difference between groups are statistically significant with a small relationship between topic proportion and each variable, respectively.

**Table 1**
*Results for the Chi-square Test of Independence with three different variables*

| Variable | Categories | Dof | N_topics | Chi-square | P-value | Cohen's omega ($\omega$) |
|---|---|---|---|---|---|---|
| year | 5 | 60 | 16 | 97.3 | 0.0017 | 0.255 |
| region | 5 | 36 | 10 | 55.6 | 0.0195 | 0.216 |
| school setting | 2 | 42 | 43 | 81.0 | 0.0003 | 0.179 |

In addition, we identified the 5 most frequent topics for each variable in their respective categories to help illustrate the relationship between topics and variables. Due to space constraints, we only present the results for the "year" (Table 2) and "region" (Table 3) variables; we used a color scheme to help identify the "intruders", i.e., topics present only in one or two intervals.

Table 2
*Top-5 topics for each year-interval explored in the study.*

| Rank | 2003-2006 | 2007-2010 | 2011-2014 | 2015-2018 | 2019-2022 |
|---|---|---|---|---|---|
| #1 | Water resources | Water resources | Waste treatment | Water resources | Sustainable agriculture |
| #2 | Robotics | Sustainable agriculture | Water resources | Waste treatment | Teaching and learning |
| #3 | Renewable energy sources | Renewable energy sources | Dengue disease | Sustainable agriculture | Water resources |
| #4 | Teaching and learning | Heritage languages | Sustainable agriculture | Dengue disease | Covid-19 |
| #5 | Oils as fuels | Learning of sustainability | Teaching and learning | Renewable energy sources | Healthy eating |

Table 3
*Top-5 topics for each region explored in the study.*

| Rank | Southeast | Northeast | South | Central West | North |
|---|---|---|---|---|---|
| #1 | Water resources | Water resources | Sustainable agriculture | Sustainable agriculture | Water resources |
| #2 | Sustainable agriculture | Dengue disease | Waste treatment | Renewable energy sources | Waste treatment |
| #3 | Teaching and learning | Waste treatment | Teaching and learning | Water resources | Sustainable agriculture |
| #4 | Renewable energy sources | Teaching and learning | Water resources | Waste treatment | Teaching and learning |
| #5 | Visual impairment | Sustainable agriculture | Renewable energy sources | Dengue disease | Traffic accidents |

Some "intruders" in Table 2 demonstrate how topics are influenced by the year in which projects were designed. For example, "dengue disease" is the seventh most common topic among all projects but became significantly more frequent in the 2010s due to an increase in cases in Brazil (Nunes et al., 2019). "Covid-19", which became relevant globally, was the fourth most common topic in 2019-2022. Similarly, in Table 3, "dengue disease", is more relevant to students from two regions (Northeast and Central West), which historically have had the highest incidence *per capita* of the disease (Catão & Guimarães, 2011).

## Discussion and future work

Our findings illustrate the diverse range of topics that students have explored in their science and engineering projects at FEBRACE (RQ1). Some of the most frequent topics relate to environmental studies (e.g., "water resources", "sustainable agriculture", "renewable energy sources") and indicate some responsiveness to the Brazil's natural resources and economy; at the same time, they support a more traditional view of scientific inquiry, associated to the Natural Sciences. On the other hand, the list of topics also provides examples from the Social Sciences, such as "teaching and learning", violence against women" and "heritage languages". Further in-depth analysis of sample projects focusing on specific topics can provide valuable insights into their relevance and potential impact on students' understanding of science inquiry and the broader scientific field. In addition, the results from statistical tests suggest the topics identified in this study are responsive to their context (RQ2), supporting their description as open inquiry processes (LaBlanca, 2012). Including additional variables, such as city size and HDI, together with the analysis of actual projects from those topics can provide a more nuanced understanding of the relationship between context and scientific inquiry.

This study has limitations stemming from the data sources used. The topic modeling analysis was based solely on projects presented at FEBRACE, as they are publicly available. Including data from all submissions could have led to different results but would still have been limited in terms of representativeness, as we cannot affirm students have equal access to the fair. Additionally, the list of topics is specific to the context of this study and may vary in other contexts. Rather than generalizing our results to any SEFs, we propose that our methodology can support the design of similar studies in different SEFs and open inquiry processes, both within and beyond Brazil. Most importantly, those findings point to the opportunity of providing students with adequate resources to support their inquiry. Open inquiry processes can be mistakenly perceived as independent of teacher guidance and support, whereas teachers play an important role in scaffolding students' inquiry (e.g., Hmelo-Silver et al., 2007). That means, for example, making sure students have access to resources that will enable their inquiry (Zion & Mendelovici, 2012), taking into account context-specific opportunities and constraints. Our results can thus inform both instruction and the design of meaningful resources that support teachers and students' inquiry not only in SEF projects, but in general open inquiry experiences.

## References

Adler, I., Schwartz, L., Madjar, N., & Zion, M. (2018). Reading between the lines: The effect of contextual factors on student motivation throughout an open inquiry process. *Science Education*, 102(4), 820-855.

Alcoforado, A., Ferraz, T. P., Gerber, R., Bustos, E., Oliveira, A. S., Veloso, B. M., ... & Costa, A. H. R. (2022). ZeroBERTo: Leveraging zero-shot text classification by topic modeling. In *International Conference on Computational Processing of the Portuguese Language* (pp. 125-136). Cham: Springer International Publishing.

Catão, R. C., & Guimarães, R. B. (2011). Mapeamento da reemergência do dengue no Brasil-1981/82-2008. *Hygeia-Revista Brasileira de Geografia Médica e da Saúde*, 7(13).

Chen, Y., Yu, B., Zhang, X., & Yu, Y. (2016, April). Topic modeling for evaluating students' reflective writing: a case study of pre-service teachers' journals. In *Proceedings of the sixth international conference on learning analytics & knowledge* (pp. 1-5).

Coelho, R. & McCollum, A. (2021). What Can Automated Analysis of Large-Scale Textual Data Teach Us about the Cultural Resources that Students Bring to Learning?. In de Vries, E., Hod, Y., & Ahn, J. (Eds.), *Proceedings of the 15th International Conference of the Learning Sciences - ICLS 2021*. (pp. 565-568). Bochum, Germany: International Society of the Learning Sciences.

Dieng, A. B., Ruiz, F. J., & Blei, D. M. (2020). Topic modeling in embedding spaces. *Transactions of the Association for Computational Linguistics*, 8, 439-453.

Egger, R., & Yu, J. (2022). A Topic Modeling Comparison Between LDA, NMF, Top2Vec, and BERTopic to Demystify Twitter Posts. *Frontiers in Sociology*, 7.

Ferraz, T. P., Alcoforado, A., Bustos, E., Oliveira, A. S., Gerber, R., Müller, N., D'Almeida, A. C., Veloso, B. M., & Costa, A. H. R. (2021). DEBACER: a method for slicing moderated debates. In *Anais do XVIII Encontro Nacional de Inteligência Artificial e Computacional* (pp. 667-678). SBC.

Grinnell, F., Dalley, S., & Reisch, J. (2020). High school science fair: Positive and negative outcomes. *PloS one*, 15(2), e0229237.

Grootendorst, M. (2022). BERTopic: Neural topic modeling with a class-based TF-IDF procedure. *arXiv preprint arXiv:2203.05794*.

Hmelo-Silver, C. E., Duncan, R. G., & Chinn, C. A. (2007). Scaffolding and achievement in problem-based and inquiry learning: A response to Kirschner, Sweller, and Clark (2006). *Educational Psychologist*, 42(2), 99–107.

Koomen, M. H., Rodriguez, E., Hoffman, A., Petersen, C., & Oberhauser, K. (2018). Authentic science with citizen science and student driven science fair projects. Science Education, 102(3), 593-644.

LaBanca, F. (2008). *Impact of problem finding on the quality of authentic open inquiry science research projects*. Western Connecticut State University.

Nunes, P. C. G., Daumas, R. P., Sánchez-Arcila, J. C., Nogueira, R. M. R., Horta, M. A. P., & Dos Santos, F. B. (2019). 30 years of fatal dengue cases in Brazil: a review. *BMC public health*, 19(1), 1-11.

Röder, M., Both, A., & Hinneburg, A. (2015, February). Exploring the space of topic coherence measures. In *Proceedings of the eighth ACM international conference on Web search and data mining* (pp. 399-408).

Zion, M., & Mendelovici, R. (2012). Moving from structured to open inquiry: Challenges and limits. *Science education international*, 23(4), 383-399.

## Acknowledgments

We thank the FEBRACE organizing committee for their support along the study. This work was partially supported by the grant Projeto Ciência na Escola - 441081/2019-3 from the Conselho Nacional de Desenvolvimento Científico e Tecnológico (CNPq) in Brazil.